\documentclass[12pt, double]{article}
\usepackage{times}
\usepackage{algorithm}
\usepackage{algorithmic}
\usepackage{wrapfig}
\usepackage{verbatim}
\usepackage{amsmath}
\usepackage{graphicx}
\usepackage{subcaption}
\usepackage{srcltx}
\usepackage{color}
\usepackage{lineno}
\usepackage{dirtytalk}
\usepackage{amsthm}
\usepackage{authblk}
\usepackage{listings}
\usepackage{tikz}
\usepackage{longtable}
\usepackage{comment}
\usetikzlibrary{shapes.geometric, arrows}
\usepackage{epsfig,amsfonts,amssymb}
\usepackage{adjustbox}
\usepackage{multirow}
\usepackage{setspace}
\usepackage{hyperref}
\usepackage{comment}
\usepackage{xcolor}

\newtheorem{definition}{Definition}

\newcommand{\review}[1]{\textcolor{black}{#1}}

\long\def\comment #1\commentend{}

\begin{document}

\title{\Large Introducing 'Inside' Out of Distribution}
\author[1,2,*]{Teddy Lazebnik}
\affil[1]{Department of Information Systems, University of Haifa, Haifa, Israel}
\affil[2]{Department of Computing, Jonkoping University, Jonkoping, Sweden}
\affil[*]{Corresponding author: lazebnik.teddy@gmail.com}
\date{}
\maketitle
\thispagestyle{empty}

\begin{abstract}
\noindent
\review{Detecting and understanding out-of-distribution (OOD) samples is crucial in machine learning (ML) to ensure reliable model performance. Current OOD studies primarily focus on extrapolatory (outside) OOD, neglecting potential cases of interpolatory (inside) OOD. In this study, we introduce a novel perspective on OOD by suggesting it can be divided into inside and outside cases. We examine the inside-outside OOD profiles of datasets and their impact on ML model performance, using normalized Root Mean Squared Error (RMSE) and \(F_1\) score as the performance metrics on syntetically-generated datasets with both inside and outside OOD. Our analysis demonstrates that different inside-outside OOD profiles lead to unique effects on ML model performance, with outside OOD generally causing greater performance degradation, on average. These findings highlight the importance of distinguishing between inside and outside OOD for developing effective counter-OOD methods.}  
\end{abstract}

\pagestyle{myheadings}
\markboth{Draft: \today}{Draft: \today}
\setcounter{page}{1}

\section{Introduction}
\label{sec:introduction}
Achieving high performance in regression and classification tasks using machine learning (ML) and deep learning (DL) models presents a fundamental computational challenge that is critical for various scientific and engineering applications \cite{virgolin2020machine,kutz2017deep,reichstein2019deep,alzubaidi2021review,ode_sr_2}
. The effectiveness of ML models depends on several factors, including the nature of the problem and the data available for training \cite{he2021automl,zhong2005comparison,discussion_1,exp_1,exp_2,li2018ease,teddy_automl,f_i_1,f_i_2}
. A growing body of research examines the characteristics of datasets in data-driven tasks, considering aspects such as noise \cite{noise_1,noise_2}
, concept drift \cite{cd_1,cd_2}
, and out-of-distribution (OOD) data \cite{ood_1,ood_2}.

The fundamental premise of data-driven models (which includes ML and DL) is based on the assumption that data will be identically and independently distributed (i.i.d) \cite{iid}. Namely, the training and test data are presumed to come from the same distribution. More importantly, the training and inference data are presumed to come from the same distribution. Nevertheless, this assumption often falls short in numerous real-world scenarios \cite{rw_t_1}. 
Over time, data-driven models have become pervasive across various domains, and their deployment in real-world settings frequently encounters violations of the i.i.d assumption \cite{hagabubim,discussion_3,meta_l_1}
. As such, it is common for these models to experience a decline in performance over time. This decline in performance is typically attributed to shifts in data distributions and a larger sample size of the real-world distribution that reveals the bias in the original training data \cite{rw_t_2,pick_tau_1}. Currently, there is an active investigation into this phenomenon, commonly referred to as "out-of-distribution" (OOD) \cite{kirchheim2022pytorch,rw_t_3}. 

OOD data and its impact on data-driven models have been extensively investigated due to its frequent occurrence and the significant challenges it presents \cite{ood_r_1,ood_r_2,ood_r_3}.
Addressing the challenges posed by OOD scenarios is crucial for ensuring the robustness and reliability of data-driven models across various applications \cite{ood_r_4,ood_r_5}.
Intuitively, OOD refers to data instances that significantly deviate from the training data distribution of data-driven models. This definition should be carefully distinguished from concept drift, which describes changes in data's underlying distribution over time. Practically speaking, OOD is a fundamental issue in data-driven modeling because it exposes the limitations of models that assume the training data is fully representative of the task's dynamics \cite{ml_assumption}. 

Indeed, multiple studies focused their efforts on tackling the challenges posed by OOD, including but not limited to OOD definition \cite{ood_definition_1,ood_definition_2}.
 OOD detection \cite{ood_detection_1,ood_detection_2,ood_detection_3}, and OOD robustness \cite{ood_better_1,ood_better_2}.
For instance, \cite{new_ood_2} builds on the Risk Extrapolation mathematical framework, employing robust optimization over a perturbation set of extrapolated domains to demonstrate that reducing differences in risk across training domains can mitigate a model’s sensitivity to a wide range of extreme distributional shifts. \cite{new_ood_3} proposed a simple mixup-based technique, called LISA, that learns invariant predictors via selective augmentation. This method selectively interpolates samples with either the same labels but different domains, or the same domain but different labels, addressing subpopulation shifts (e.g., imbalanced data) and domain shifts. Moreover, \cite{new_ood_4} extended the task of improving robustness to OOD by incorporating an OOD detection mechanism as an integral part of the method. Specifically, the authors propose a margin-based learning framework that exploits freely available unlabeled data in the wild, capturing the environmental test-time OOD distributions under both covariate and semantic shifts. Additionally, \cite{new_ood_1} empirically showed that OOD performance is strongly correlated with in-distribution performance across a wide range of models and distribution shifts. They connected the strength of this correlation to the Gaussian data model, revealing that the further a sample is from the Gaussian-defined centroid, the weaker the correlation becomes.

\begin{figure}[!ht]
    \centering
    \includegraphics[width=0.99\textwidth]{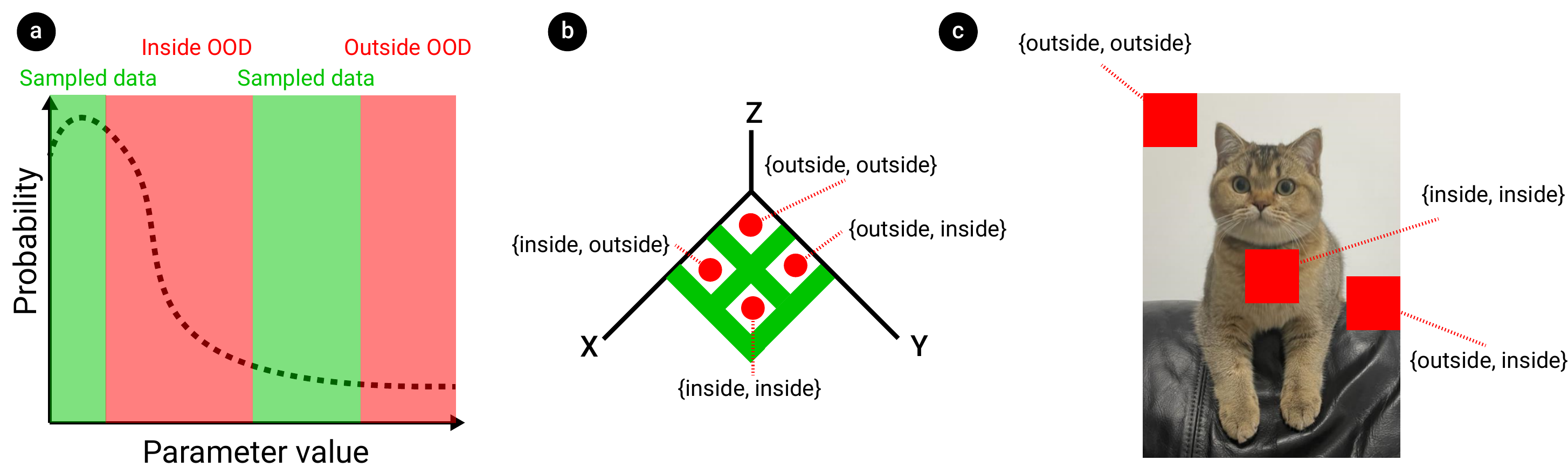}
    \caption{An example of inside and outside out of distribution (OOD) for (a) one, (b) two, and (c) five dimensions.}
    \label{fig:intro}
\end{figure}

Notably, these attempts mainly focused on OOD as an outlier phenomenon for a given dataset, ignoring the cases where OOD is \textbf{surrounded} by the available (i.e., training) data. With this perspective, one can consider a binary separation between two types of OOD - \say{inside} and \say{outside} of the training data. Fig. \ref{fig:intro} provides several examples of these differences such that (a) shows inside and outside OOD for one-dimensional case; (b) shows the four possible inside and outside OOD configurations possible in two-dimensional case; and (c) shows inside and outside OOD for the task of in-painting from computer vision which defined in a five-dimensional case corresponding to the three color dimensions and the two location dimensions \cite{inpainting}.

\review{In this study, we investigate the OOD phenomenon from the inside-outside perspectives. We start by defining the inside-outside OOD profile of a \(n\)-dimensional distribution, followed by numerically profiling it for a large number of synthetic datasets. In addition, we perform a sensitivity analysis to draw rules of thumb for practical guidelines for practitioners tackling complex OOD profiles.}

The rest of this manuscript is organized as follows. Section \ref{sec:related_work} provides an overview of OOD definitions. Section \ref{sec:model} formally introduces the OOD inside and outside OOD definitions. Section \ref{sec:numerical_analysis} outlines the experimental setup used as well as the numerical results obtained. Finally, in section \ref{sec:discussion}, we discuss the applicative outcome of the obtained results and suggest possible future work.  

\section{Related Work}
\label{sec:related_work}
\begin{figure}[!ht]
    \centering
    \includegraphics[width=0.6\textwidth]{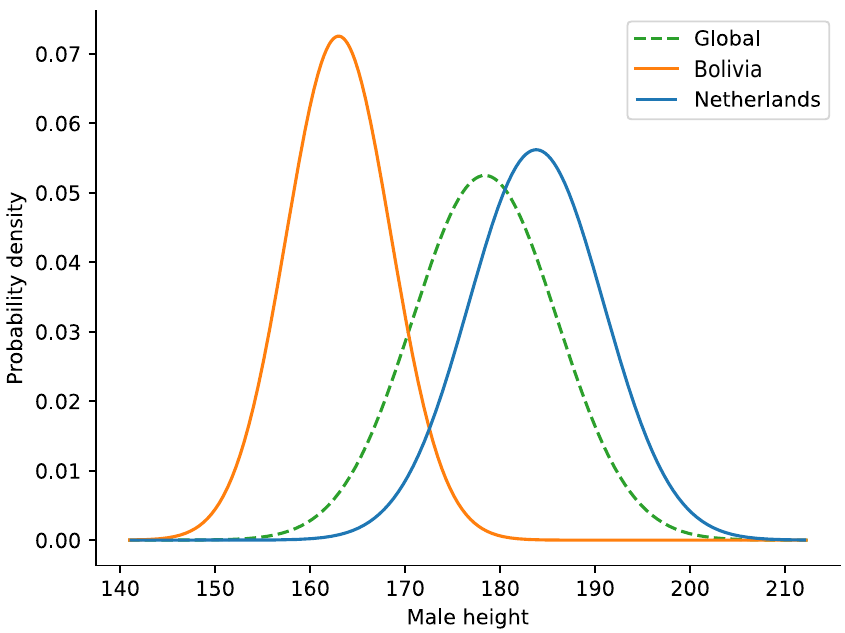}
    \caption{An example for inside out of distribution (OOD) for the case of males' height distribution given sampled data from one of the shortest (Bolivia) and highest (Netherlands) represented countries. }
    \label{fig:related_work}
\end{figure}

OOD is a complex mathematical phenomenon as it emerges from gaps between the information available to an observer and the real-world dynamics occurring in practice. To illustrate this idea, let us consider a simple case of height distribution in the male population, which is known to be distributed normally with a mean value of 178.4 centimeters and a standard deviation of 7.6 centimeters \cite{human_height}. 
A researcher collects data from Bolivia and the Netherlands with an average male height of 163.0 (SD = 5.5) centimeters \cite{human_height_2}
and 183.8 centimeters (SD = 7.1) \cite{human_height_1}, respectively. In such a case, if the researcher used ML to produce a model that predicts the height of males, it would obtain poor results for males from most countries, as these are not well-represented in the data. Due to the fact that Bolivia and the Netherlands have the shortest and highest male populations, respectively, the OOD is between them and therefore inside. Fig. \ref{fig:related_work} visualizes this example. 

As such, multiple definitions have been proposed for OOD over the years \cite{ood_def}. 
From these, three well-adopted OOD definitions are the Mahalanobis distance, random-feature Z-score, and Kullback–Leibler (KL) divergence metric \cite{kirchheim2022pytorch}. Mahalanobis distance is an extension of the Z-score metric for \(n\)-dimensional case \cite{Mahalanobis_2018}. Formally, considering a probability distribution $Q$ over $\mathbb{R}^N$, characterized by its mean vector $\vec{\mu} = (\mu_1, \mu_2, \mu_3, \ldots, \mu_N)^{\mathsf{T}}$ and a positive-definite covariance matrix $S$, the Mahalanobis distance $d_M$ from a point $\vec{x} = (x_1, x_2, x_3, \ldots, x_N)^{\mathsf{T}}$ to $Q$ is defined as:\[ d_M(\vec{x}, Q) = \sqrt{(\vec{x} - \vec{\mu})^{\mathsf{T}}S^{-1}(\vec{x} - \vec{\mu})}.\] This definition of the multivariate Z-score has been extensively used and is likely the most common for OOD in regression tasks \cite{Bendale_et_al,Lee_et_al,Mayrhofer_Filzmoser_2023}. A random-feature Z-score selects at random one feature in each iteration and determines whether the observation is in or OOD based on its Z-score, given some threshold \cite{new_ood_5}. KL divergence, a concept from information theory, measures how one probability distribution diverges from a second, reference probability distribution. In the context of OOD, KL divergence is used to quantify the difference between the probability distributions of in-distribution and OOD data \cite{new_ood_6}. 

\textbf{Beyond these standard methods, \cite{editor_1} propose a new measure of discrepancy specifically designed for comparing multivariate distributions, which may offer a different perspective on OOD detection by focusing on overall distributional differences. Another approach to addressing the challenge of measuring inequality in multivariate distributions involves extending the univariate Gini coefficient \cite{editor_3}. Such approaches often aim to maintain desirable properties of the Gini coefficient while adapting it to higher-dimensional data, frequently employing techniques like whitening processes to ensure scale stability. Moreover, \cite{editor_2} proposes novel multidimensional inequality indices, unlike traditional indices that often leverage techniques like the Fourier transform \cite{bracewell1989fourier} to create scaling-invariant measures that are more computationally tractable than traditional methods like Lorenz Zonoids \cite{koshevoy1996lorenz}. The authors quantify mutual variability rather than just deviations from the mean, and often produce indices easily interpretable in a multidimensional setting, which can be used as anomaly detection metrics. \cite{itai2025tighten} leveraged the convex hull property of a dataset and the fact that anomalies highly contribute to the increase of the convex hull's volume to propose an anomaly detection algorithm which computes the convex hull's volume as an increasing number of data points are removed from the dataset to define a decision line between OOD and in-distribution data points. } 

\textbf{Moreover, ML-based anomaly detection algorithms have been proposed \cite{nassif2021machine}. For instance, Isolation Forest \cite{cheng2019outlier} leverages recursive data partitioning to isolate individual points, effectively flagging anomalies that are sparse and easily separated, particularly in high-dimensional spaces, though it struggles with densely clustered or geometrically complex anomalies due to its reliance on random splits and ordinal variable ranking. Single-Class Support Vector Machines \cite{oza2018one}, conversely, learn a boundary around normal data, classifying points outside this learned boundary as anomalies, operating under the assumption that points outside the learned boundary are anomalous. Gaussian Mixture Models \cite{li2016anomaly} model data as a mixture of Gaussian distributions, identifying anomalies based on low likelihood, a strategy effective for approximately Gaussian-shaped clusters but less suitable for data with non-Gaussian clusters or complex distributions.}

Nonetheless, these methods are not able to capture inside OOD as they take into consideration the entire training dataset and are symmetric in terms of the anomaly compared to the center of mass of the training dataset. 

\section{Out Of Distribution Profile}
\label{sec:model}
The OOD inside-outside profile aims to capture the performance of a data-driven model on the different possible OOD configurations of a given problem. Formally, let us consider a \(n+1\)-dimensional regression task where \(n\) dimensions are the source features (\(x\)) and the \(n+1\) dimensional is the target feature (\(y\)). In addition, let us assume for each dimension, \(i \in [0, \dots, n + 1]\), a sampling distribution is available and denoted by \(s_i\). The dataset (\(D = [f_1, \dots, f_{n+1}]\)) is therefore constructed by sampling \(s_i\) for \(i \in [0, \dots, n + 1]\) for \(k\) times. 

Based on this configuration, let us assume an OOD predictor, \(O\), which accepts \(\forall i \in [0, \dots, n+1]: f_i\) and a new sample \(x_i \in \mathbb{R}\) and returns either this sample is OOD with respect to \(f_i\) or not. On top of that, we would like to check if the OOD is \say{outside} or \say{inside} with respect to \(f_i\). As such, we define \textit{inside} OOD as follows:
\begin{definition}
A sample \(x\) is said to be \say{inside} OOD in the \(i_{th}\) dimension with respect to a dataset (\(D\)) if and only if \(x\) is OOD with respect to \(f_i\) given a predictor \(O\) and \(\exists v_1, v_2 \in f_i: v_1 < x < v_2\). 
\end{definition}
In a complementary manner, one can define \textit{outside} OOD as follows:
\begin{definition}
A sample \(x\) is said to be \say{outside} OOD in the \(i_{th}\) dimension with respect to a dataset (\(D\)) if and only if \(x\) is OOD with respect to \(f_i\) given a predictor \(O\) and \(!\exists v_1, v_2 \in f_i: v_1 < x < v_2\). 
\end{definition}
Based on these two definitions, the OOD profile of a dataset for a given new sample \(x\) is denoted by \(p(D, x)_O \in \{no, inside, outside\}^{n}\). 

These definitions are computationally appealing, as one is not required to solve any complex binary predicate on top of the OOD predicator, \(O\), as finding if the condition \(\exists v_1, v_2 \in f_i: v_1 < x < v_2\) is met is \(O(N)\) where \(N\) is the size of the training set without pre-process and even \(O(1)\) if the largest and smallest values of each feature are pre-computed. That said, they are also sensitive to outliers. For example, let us assume a dataset with 1000 samples, 999 out of them range between 0 and 1 while one sample is located in an arbitrary large value \(M >> 1\). In such a case, a sample with a value of 10 will be considered \textit{inside} OOD as \(v_1 = 1\) and \(v_2 = M\) satisfies \(1 \leq 10 \leq M\). However, this is clearly not the behavior one would desire for the \textit{inside} OOD definition. As such, the \textit{inside} OOD can be defined also as follows:
\begin{definition}
A sample \(x\) is said to be \say{outside} OOD in the \(i_{th}\) dimension with respect to a dataset (\(D\)) if and only if there are \(D_l, D_r \subset D \wedge D_l \cap D_r = \emptyset \) such that \(x\) is OOD with respect to \(D_l(f_i)\) and \(D_r(f_i)\) given a predictor \(O\) and \(\exists v_1 \in D_l, v_2 \in D_r: v_1 < x < v_2\). 
\end{definition}
This definition is more robust, while also more computationally extensive, as one is required to ensure there is no division of the dataset \(D\) into two subsets (\(D_l, D_r\)) that satisfy the \textit{inside} OOD. In the rest of the paper, we utilized the first definition due to its simplicity, and since the synthetic data generation procedure used does not generate anomalies, which can cause bias in the results.

\section{Numerical Analysis}
\label{sec:numerical_analysis}
In this section, we investigate the behavior of OOD inside-outside profile influence on machine learning (ML) model performance. To this end, we first define an experimental setup where the synthetic data generation procedure allows us to obtain all OOD inside-outside configurations. 

\subsection{Experimental setup}
In order to explore OOD inside-outside profiles, one needs a dataset generation procedure that allows one to obtain both inside and outside OOD for each of the dimensions while also providing a meaningful regression task that represents as many realistic datasets as possible. Hence, we divide the dataset generation task into two steps - source features generation and target feature generation.   

The source features generation generates each feature independently, so the number of features (i.e., \(n\)) does not play a part in this step. For each feature, we first generate a distribution, \(d\), which is a sum of a set of pre-defined distributions \(\{d^1, d^2, \dots d^N\}\) such that the number of contracted distributions, \(z\), is picked at random to be an arbitrary positive integer and we allow the same contracted distributions to be picked multiple times. Using \(d\), four values (\(a_1, b_1, a_2, b_2\)) indicating the start and end of the observable values of a random variable \(\eta \sim d\) such that \(a_1 < b_1 < a_2 < b_2\) are picked. The distribution \(d\) is sampled at random such that if the obtained value \(c\) satisfies \(a_1 \leq c \leq b \wedge a_2 \leq c \leq b_2\), the value is added to \(f\) and ignored otherwise. This process repeats until \(k\) samples are added to \(f\). Next, for the second step, a random function described by a symbolic regression expression tree is made with all the features in the dataset and produces the target feature. To ensure that an ML model would just reconstruct this function, a random Gaussian noise with a mean equal to zero and a standard deviation larger than zero is added to each source feature and to the target feature.

Importantly, for the OOD predictor (\(O\)), we used the K-nearest neighbors (KNN) algorithm \cite{knn} with a threshold \(\chi > 0\). The value of \(\chi\) is set to be the diameter of the largest centroid fitted on the training data using the X-means algorithm \cite{xmeans}. \review{This way, samples with an average distance of more than \(\chi\) do not belong to any cluster present in the training data, which is OOD. However, the KNN algorithm also allows us to obtain an inside OOD since a sample can be outside of the radius from two clusters but still between them, as desired.} 

Table \ref{table:hyperparameters} presents the hyperparameters used by the dataset generation procedure with their descriptions and value range. \review{The value ranges in the tables are taken to represent realistic dataset sizes and value ranges \cite{params_1,params_2,params_3} while also balancing with computational time.} 

\begin{table}[h!]
\centering
\label{tab:hyperparameters}
\begin{tabular}{p{0.2\textwidth}p{0.45\textwidth}p{0.25\textwidth}}
\hline \hline
\textbf{Hyperparameter}  & \textbf{Description}& \textbf{Value Range} \\ \hline \hline
$n$ & Number of source features  & $[1-50]$  \\
$k$ & Number of samples  & $[100-100000]$  \\
$s_i$ & Sampling distribution for each dimension $i$ & \([Gaussian(\mu \in [-5, 5], \sigma \in [-10, 10]), Uniform(a<b \in [-10, 10]), Exponential(\lambda \in [0, 10]), Weibull(\lambda \in [0, 5], k \in[ 1,5]), Beta(\alpha \in [0, 5], \beta \in [0, 5])]\) \\
$z_i$& Number of contracted distributions (from \(s_i\)) & $[1-20]$\\
$a_1, b_1, a_2, b_2$  & Start and end of observable values of a random variable generating \(f_1\) & [-100, 100] \\
$O$& OOD predictor function  & KNN \cite{knn} \\
\(Noise(0, \sigma)\) & Noise added to each source feature and target feature to prevent the ML model from simply reconstructing the function & $\sigma \in (0.1, 1)$ \\ 
\(\tau\) & \review{The number of classes for classification tasks} & [2-7] \\ 
\(\zeta\) & \review{The number of dividing points in the target feature for classification tasks} & [1-10] \\ 
Symbolic regression express tree & Function used to generate the target feature  & SciMed \cite{scimed} \\  
TPOT's number of generation & The number of iterations for the automatic ML tool & 20 \\ 
TPOT's population size & The number of ML pipelines in each iteration of the automatic ML tool & 50 \\ \hline \hline
\end{tabular}
\caption{Hyperparameters for OOD inside-outside profile experiments.}
\label{table:hyperparameters}
\end{table}

As the obtained tasks are regression and sensitive to both scaling and the dataset's dimension, we used the Root Mean Squared Error (RMSE) metric normalized to the average RMSE value where the samples are in-distribution \cite{shmuel2024symbolic}. The metric performance for all OOD profile configurations is computed to be \(n=100\) random samples from the distribution \(D\). In order to obtain an arguably best ML model for each dataset, we utilize the Tree-based Pipeline Optimization Tool (TPOT) automatic machine learning \cite{tpot} due to its effectiveness in finding near-optimal ML models for a wide range of tabular tasks \cite{substract,tpot_example_1,tpot_example_2,tpot_example_3}. 

\review{Moreover, in order to also evaluate classification tasks, for any given regression task, we randomly picked a number of classes (\(\tau \in \mathbf{N}\)) and number of dividing points for the range of the target feature (\(\zeta \in \mathbf{N} > \tau -2\)). For each range, divided by either two division points or a division point and the edge of the target feature's distribution, we assign a value of a class at random, allowing repetitions only after each class is assigned at least once. In order to evaluate the classification tasks, we adopted the \(F_1\) score  \cite{fourure2021anomaly}, which is commonly used for anomaly detection tasks.} 

\subsection{Results}
The results of the above analysis are divided into two parts - profiling of the inside-outside distribution for the average case for different dimensions and the sensitivity of ML models' performance as properties of the inside-outside OOD change.

\subsubsection{Profiling}
\begin{figure}[!ht]
    \centering
    \begin{subfigure}{.32\textwidth}
        \includegraphics[width=0.99\textwidth]{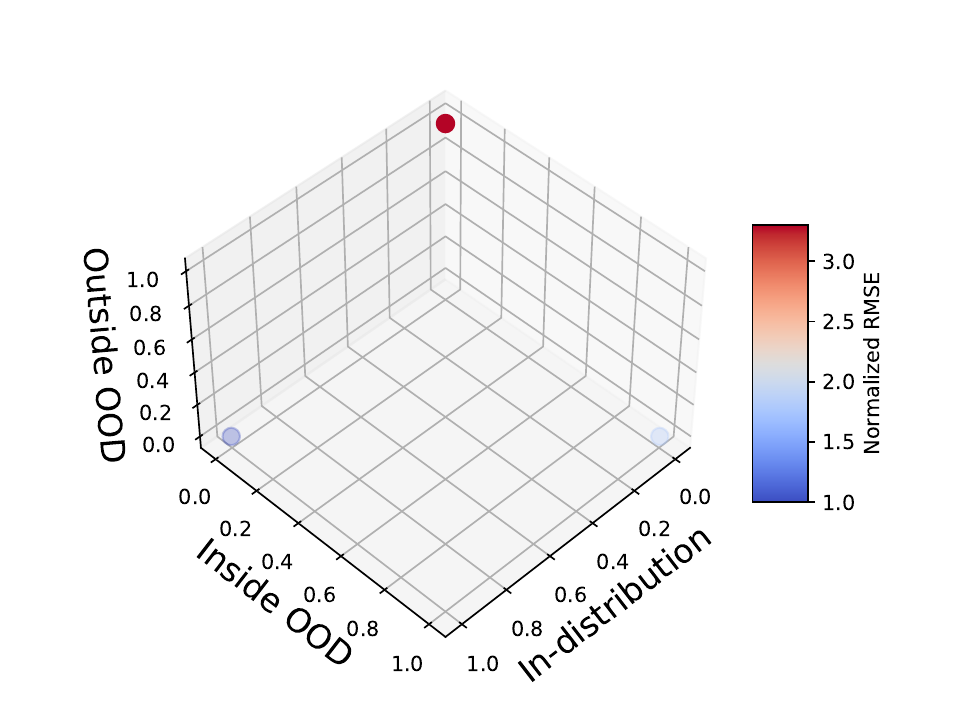}
        \caption{1-dimensional.}
        \label{fig:d1}
    \end{subfigure}   
    \begin{subfigure}{.32\textwidth}
        \includegraphics[width=0.99\textwidth]{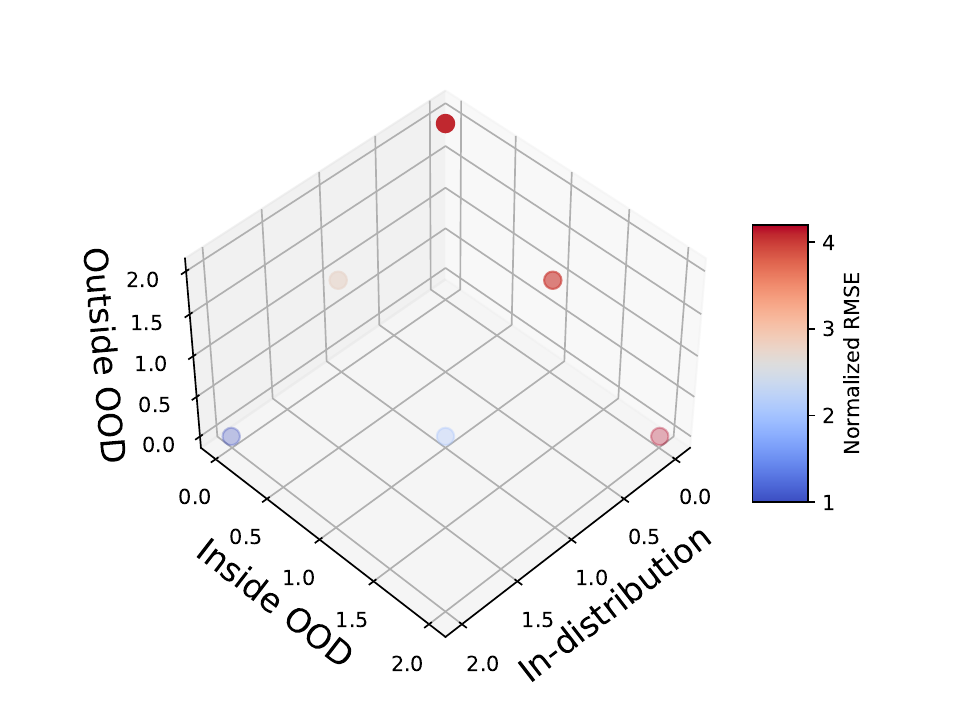}
        \caption{2-dimensional.}
        \label{fig:d2}
    \end{subfigure}   
    \begin{subfigure}{.32\textwidth}
        \includegraphics[width=0.99\textwidth]{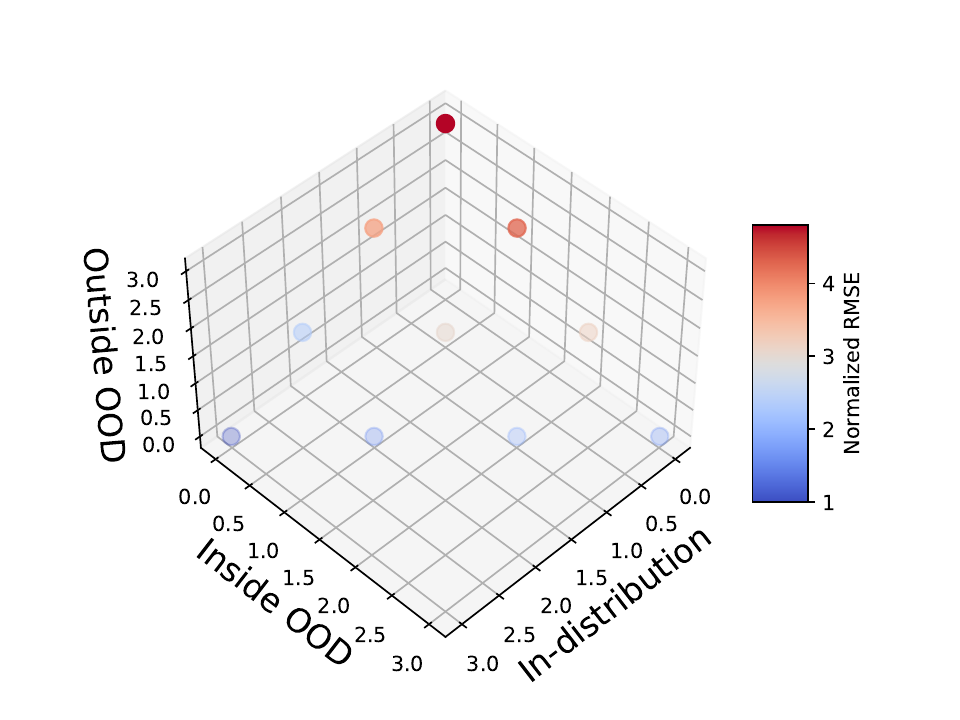}
        \caption{3-dimensional.}
        \label{fig:d3}
    \end{subfigure} 
    
    \begin{subfigure}{.99\textwidth}
        \includegraphics[width=0.99\textwidth]{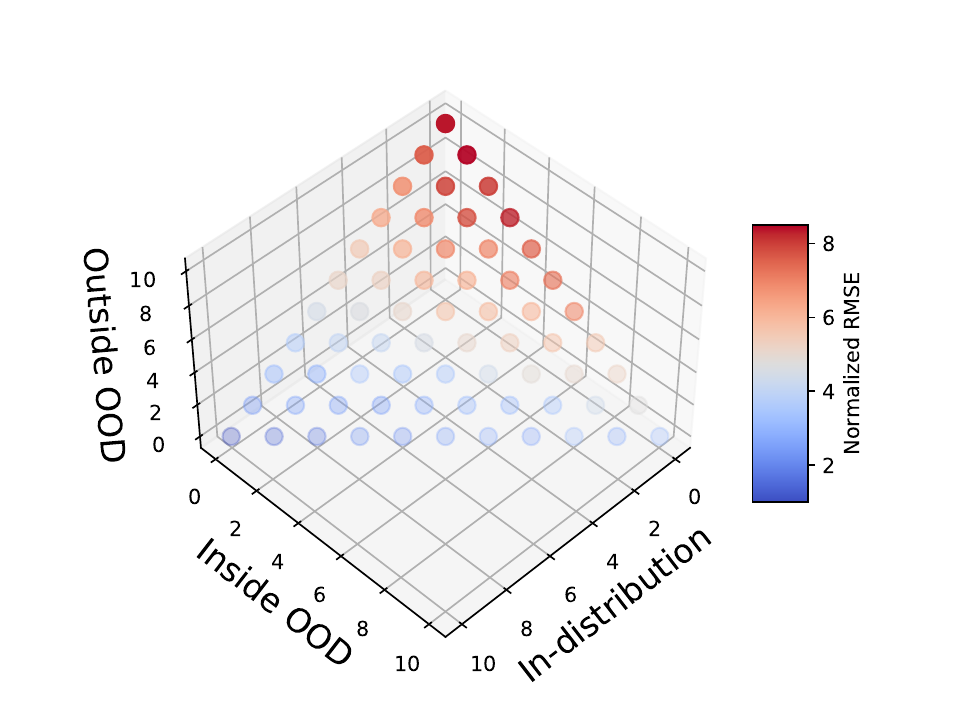}
        \caption{10-dimensional.}
        \label{fig:d10}
    \end{subfigure} 
    
    \caption{The inside-outside out of distribution (OOD) profiles (\(p(D, x)_O\)) for different dimensions \review{for regression tasks}. The color indicates the average normalized RMSE for \(n=100\) repetitions.}
    \label{fig:profile}
\end{figure}

Figure \ref{fig:profile} presents the inside-outside OOD profiles (\(p(D, x)_O\)) for different dimensions. The x,y, and z axis are corresponding to the in-distribution, inside OOD, and outside OOD, respectively. The dots are located as the number of features associated with each type as the features are symmetric in our context, the order is not important. The color, ranging from blue to red indicates the average normalized RMSE for \(n=100\) repetitions. From the sub-figures \ref{fig:d1}, \ref{fig:d2}, and \ref{fig:d3}, indicating all possible inside-outside OOD configurations for 1, 2, and 3 dimensions, respectively, a pattern that outside OOD results in higher RMSE and the effect is non-linear to the number of dimensions. Extended to a more realistic configuration, sub-figure \ref{fig:d10} shows the same analysis for \(n=10\) dimensions. The pattern that emerged from the first three dimensions is preserved. 

\begin{figure}[!ht]
    \centering
    \begin{subfigure}{.32\textwidth}
        \includegraphics[width=0.99\textwidth]{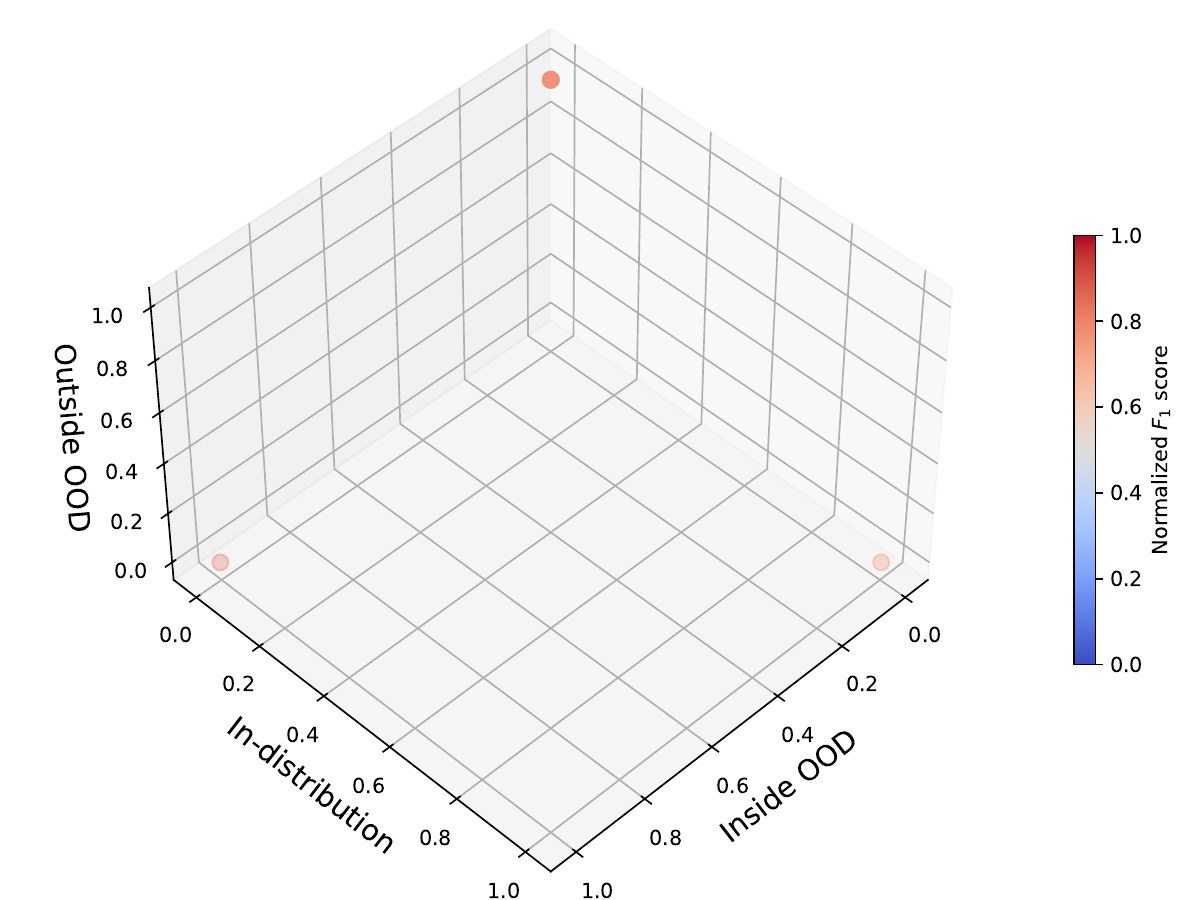}
        \caption{1-dimensional.}
        \label{fig:class_d1}
    \end{subfigure}   
    \begin{subfigure}{.32\textwidth}
        \includegraphics[width=0.99\textwidth]{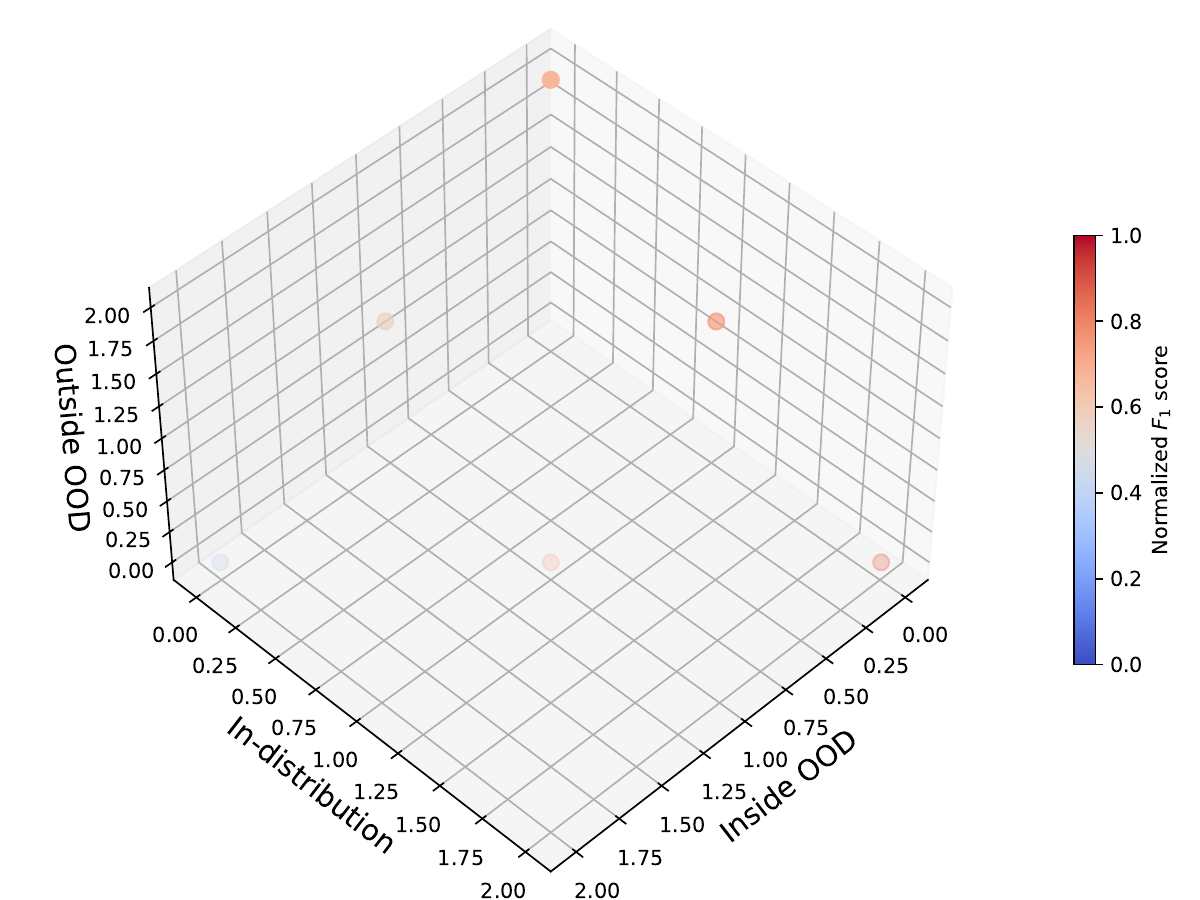}
        \caption{2-dimensional.}
        \label{fig:class_d2}
    \end{subfigure}   
    \begin{subfigure}{.32\textwidth}
        \includegraphics[width=0.99\textwidth]{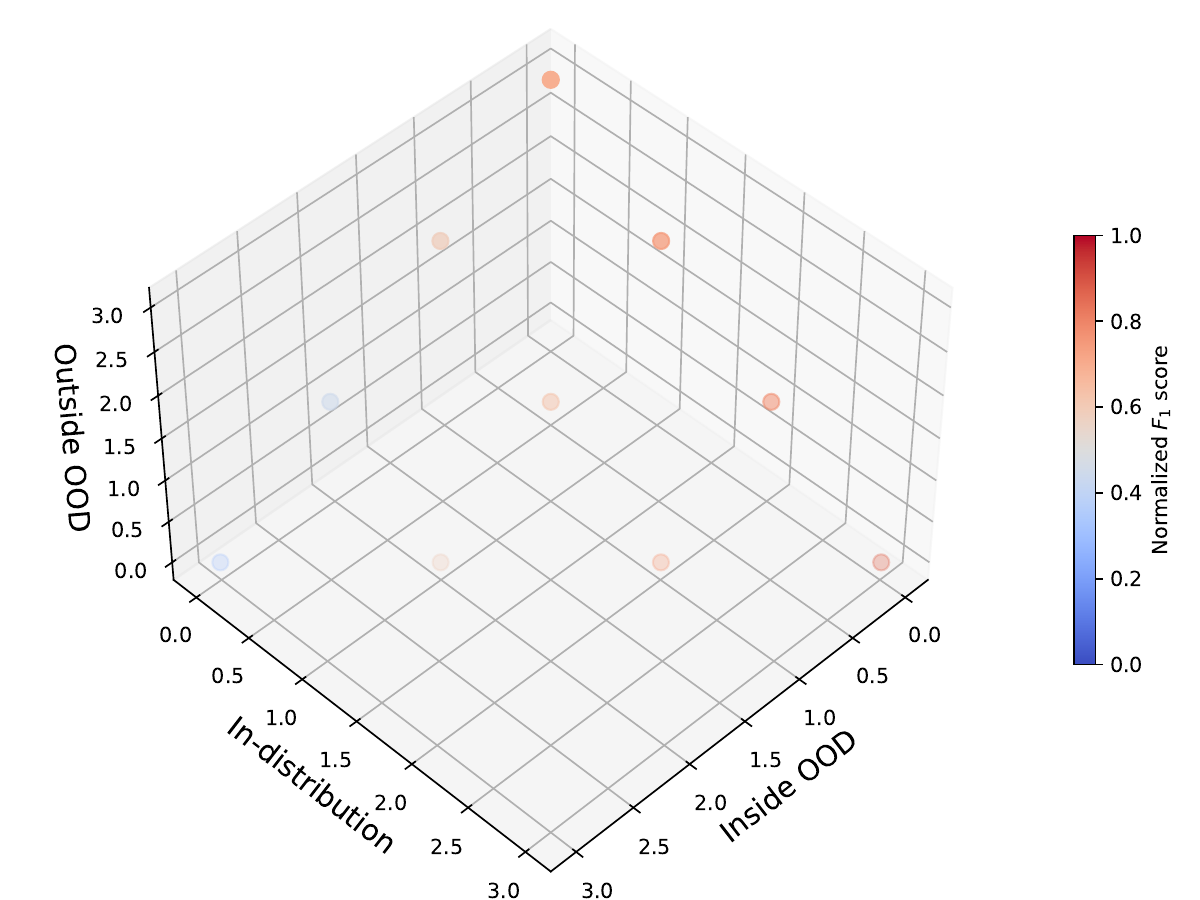}
        \caption{3-dimensional.}
        \label{fig:class_d3}
    \end{subfigure} 
    
    \begin{subfigure}{.99\textwidth}
        \includegraphics[width=0.99\textwidth]{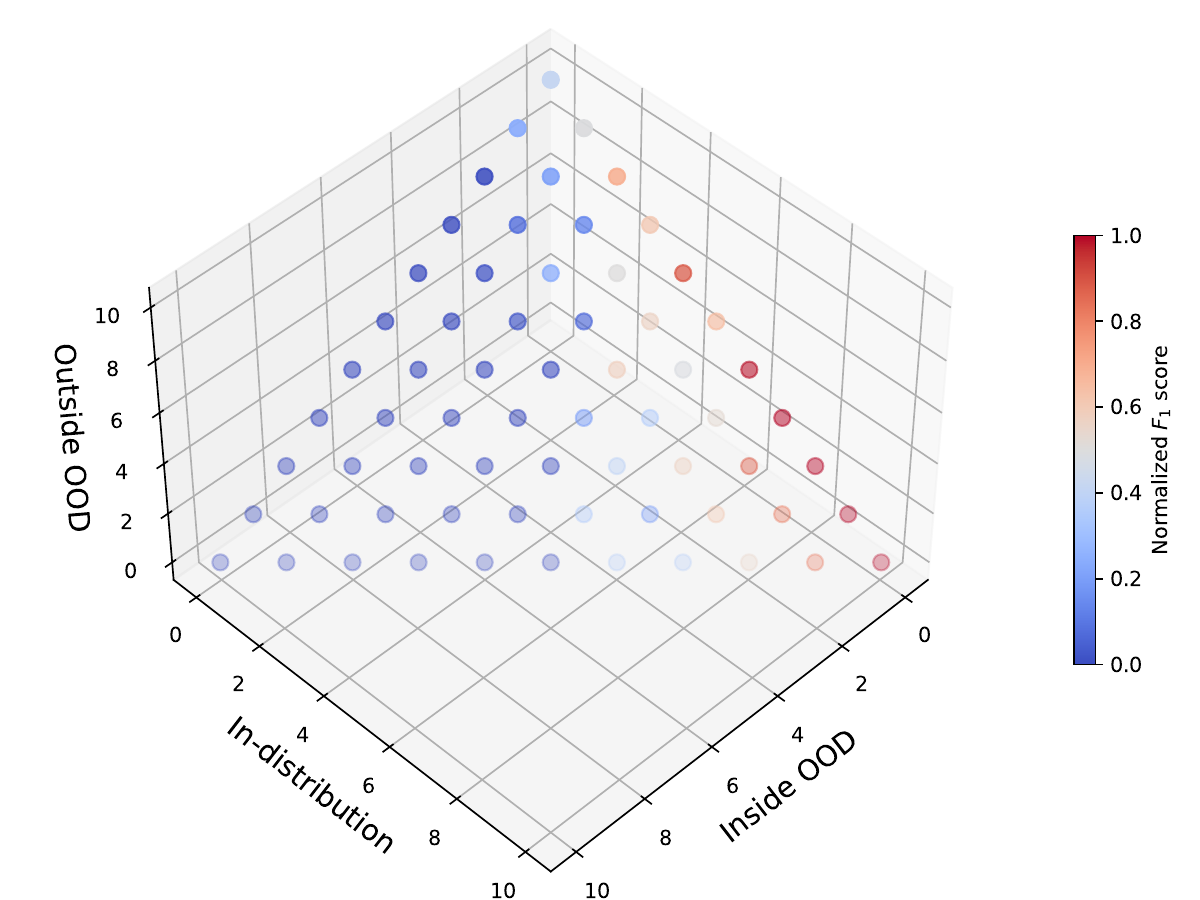}
        \caption{10-dimensional.}
        \label{fig:class_d10}
    \end{subfigure} 
    
    \caption{The inside-outside out of distribution (OOD) profiles (\(p(D, x)_O\)) for different dimensions \review{for classification tasks}. The color indicates the average normalized RMSE for \(n=100\) repetitions.}
    \label{fig:class_profile}
\end{figure}

\review{Figure \ref{fig:class_profile} presents the inside-outside OOD profiles (\(p(D, x)_O\)) for different dimensions for synthetic classification tasks. Like Figure \ref{fig:profile}, the x,y, and z axis are corresponding to the in-distribution, inside OOD, and outside OOD, respectively, and the dots are located as the number of features associated with each type while the color, ranging from blue to red indicates the average normalized \(F_1\) score for \(n=100\) repetitions. The sub-figures \ref{fig:class_d1}, \ref{fig:class_d2}, and \ref{fig:class_d3}, showcasing inside-outside OOD configurations for 1-3 dimensions, indicate that outside OOD consistently produces lower \(F_1\) score values, and this effect grows non-linearly with dimensionality. This pattern is maintained when extended to \(n=10\), as shown in sub-figure \ref{fig:d10}. In particular, one can notice that a combination of inside OOD and outside OOD results in lower \(F_1\), in comparison to only inside or outside OOD.}

\subsubsection{Sensitivity analysis}
\begin{figure}[!ht]
    \centering
    \begin{subfigure}{.49\textwidth}
        \includegraphics[width=0.99\textwidth]{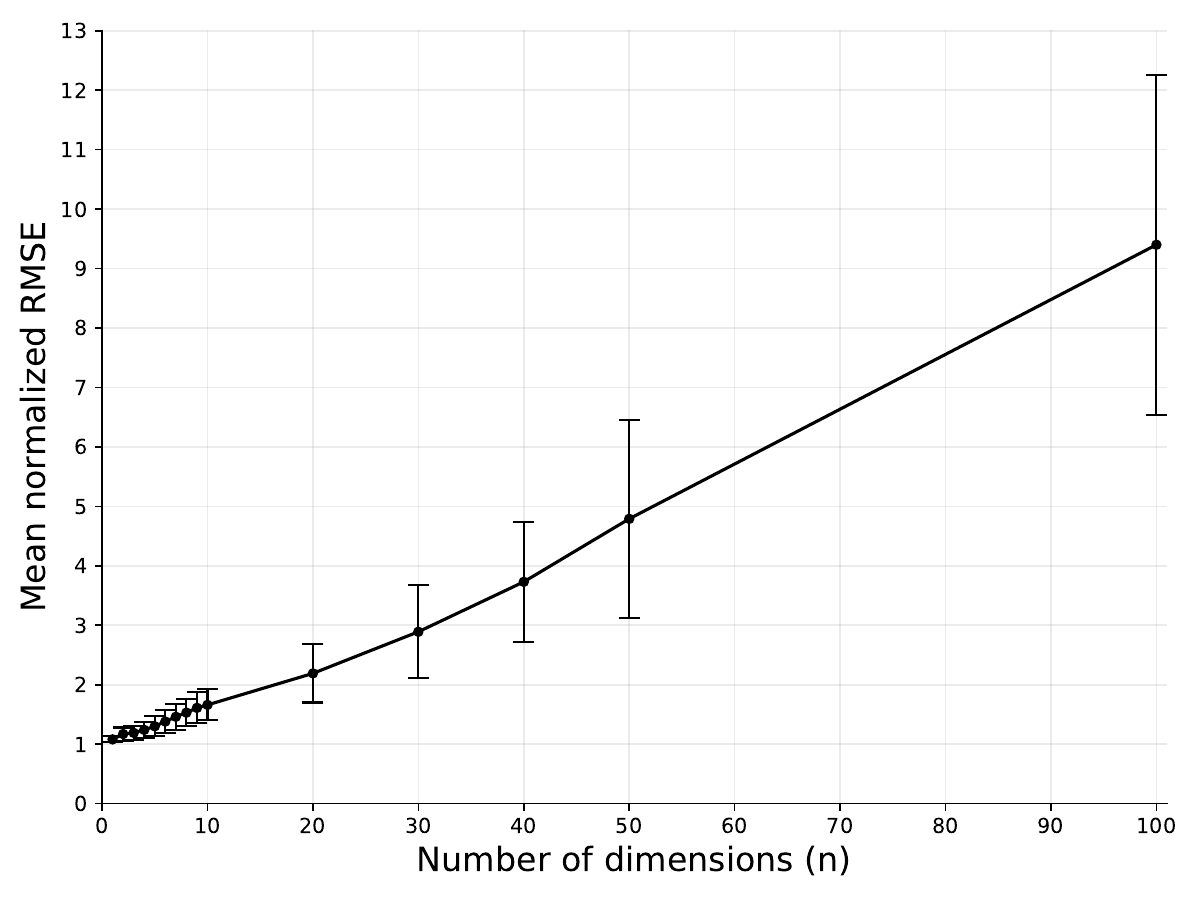}
        \caption{Number of dimensions (\(n\)).}
        \label{fig:sens_dim}
    \end{subfigure}   
    \begin{subfigure}{.49\textwidth}
        \includegraphics[width=0.99\textwidth]{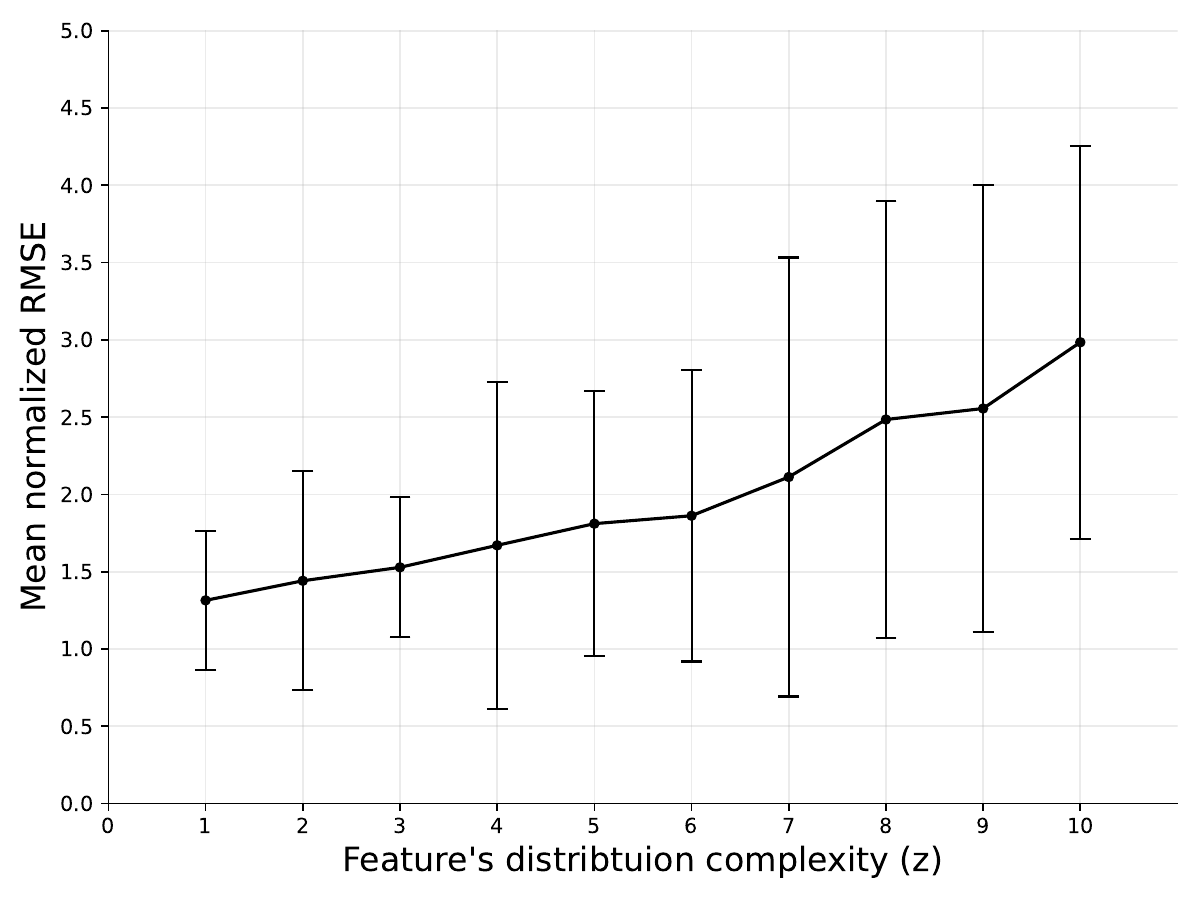}
        \caption{Feature's distribution complexity (\(z\)).}
        \label{fig:sens_complexity}
    \end{subfigure}   
    
    \caption{A one-dimensional sensitivity analysis of inside-outside out of distribution (OOD). The results are shown as the mean \(\pm\) standard deviation of \(n=100\) repetitions.}
    \label{fig:sens_1d}
\end{figure}

Figures \ref{fig:sens_dim} and \ref{fig:sens_complexity} present the mean normalized RMSE as a function of the number of dimensions and the feature's distribution complexity, respectively. The results are shown as the mean \(\pm\) standard deviation of \(n=100\) repetitions. Both sub-figures present a one-dimensional sensitivity analysis of inside-outside OOD with changes in the complexity of the regression task. sub-figure \ref{fig:sens_dim} shows a semi-linear growth in the mean normalized RMSE with respect to the number of dimensions. At the same time, the distribution of the RMSE over the cases also increases with the dimension of the task, as indicated by the increase in the error bars. Similarly, the mean normalized RMSE monotonically increases with the feature's distribution complexity, while the error bar sizes do not show a clear pattern. This outcome can be explained as different core distribution additions can result in a much more computationally expressive distribution compared to others (for example, a sum of normal distributions compared to a sum of all unique distributions), which results in a complex pattern. 
 
\begin{figure}[!ht]
    \centering
    \begin{subfigure}{.49\textwidth}
        \includegraphics[width=0.99\textwidth]{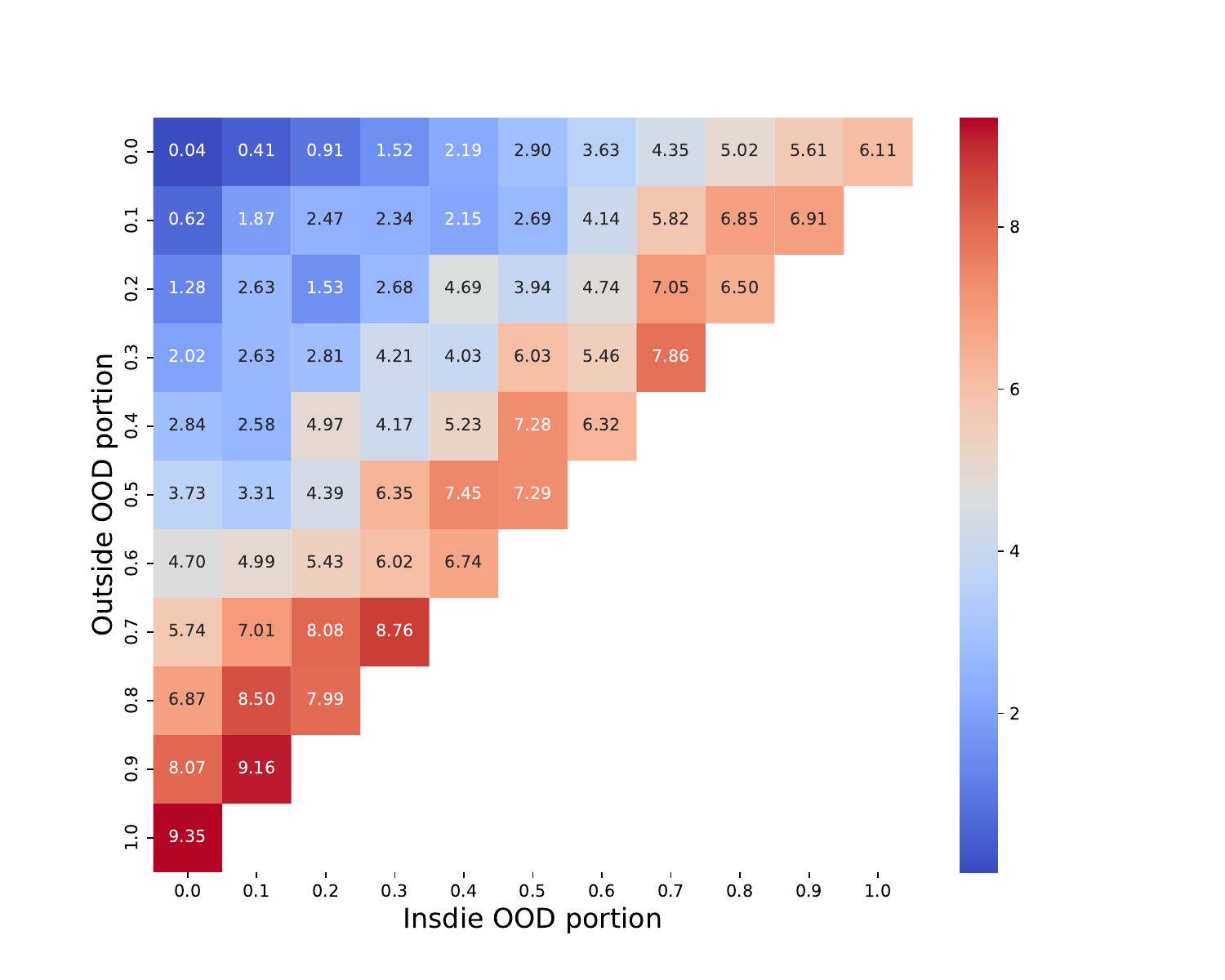}
        \caption{OOD inside-outside portion.}
        \label{fig:sens_ood_portion}
    \end{subfigure}   
    \begin{subfigure}{.49\textwidth}
        \includegraphics[width=0.99\textwidth]{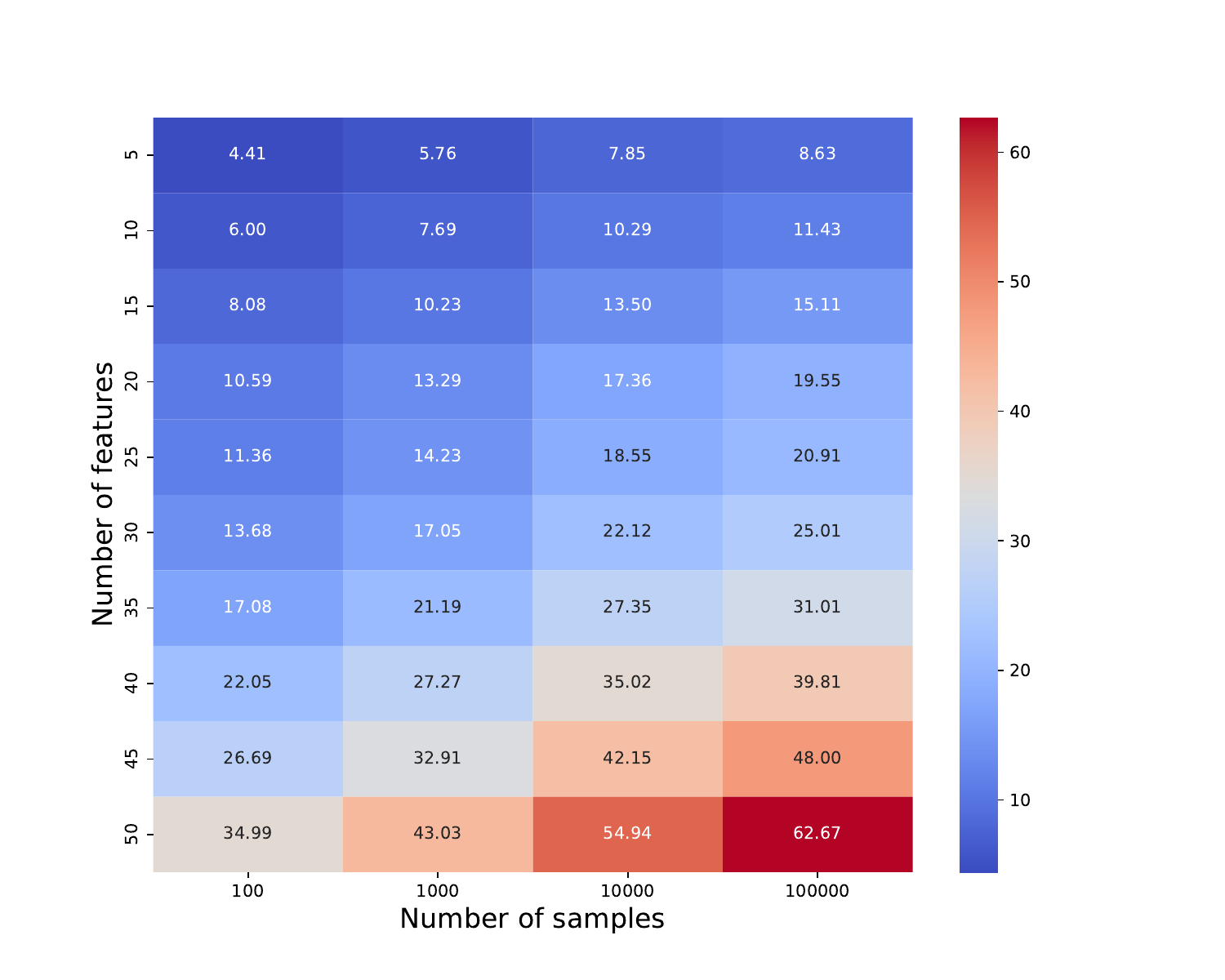}
        \caption{Dataset size.}
        \label{fig:sens_ood_size}
    \end{subfigure}   
    
    \caption{A two-dimensional sensitivity analysis of inside-outside out of distribution (OOD). The results are shown as the mean \(\pm\) standard deviation of \(n=100\) repetitions. }
    \label{fig:sens_2d}
\end{figure}

Figures \ref{fig:sens_ood_portion} and \ref{fig:sens_ood_size} present the mean normalized RMSE as a function of the inside to outside OOD portions and the dataset's size, respectively. The results are shown as the mean \(\pm\) standard deviation of \(n=100\) repetitions. Both sub-figures present a two-dimensional sensitivity analysis of inside-outside OOD with changes in the complexity of the regression task. sub-figure \ref{fig:sens_ood_size} shows that an overall increase in OOD results in higher average RMSE, with the outside OOD increasing the RMSE more than the inside OOD, aligning with the pattern presented in figure \ref{fig:profile}. In a complementary manner, sub-figure \ref{fig:sens_ood_size} reveals that an increase in the number of features and samples increases the average RMSE such that the increase in the features is semi-linear, as also indicated by sub-figure \ref{fig:sens_dim}. 

\section{Discussion}
\label{sec:discussion}
In this study, we proposed a novel perspective on OOD in the form of both inside and outside OOD, focusing on how different inside-outside OOD configurations affect the performance of ML models. Based on the inside-outside OOD definition, we conducted both profiling and sensitivity analysis on synthetic datasets for regression tasks solved using ML models. 

The results, illustrated in figures \ref{fig:profile}-\ref{fig:sens_2d}, highlight the impact of the inside-outside OOD profiles on model performance. Specifically, we show that outside OOD configurations consistently lead to higher normalized RMSE compared to inside OOD configurations across different dimensions (figures \ref{fig:profile} and \ref{fig:class_profile}). This trend is evident in one-dimensional (Figures \ref{fig:d1} and \ref{fig:class_d1}), two-dimensional (Figures \ref{fig:d2} and \ref{fig:class_d2}), three-dimensional (Figures \ref{fig:d3} and \ref{fig:class_d3}), and ten-dimensional (Figures \ref{fig:d10} and \ref{fig:class_d10}) datasets, indicating a robust pattern where outside OOD significantly degrades performance. Furthermore, the sensitivity analysis reveals a semi-linear increase in mean normalized RMSE \review{(and semi-linear decrease in the \(F_1\) score} with the number of dimensions and feature distribution complexity, which aligns with previous OOD studies \cite{editor_1}. This outcome indicates that in terms of error profile, inside and outside OOD behave similarly \cite{rw_t_3,last_1}. 
These findings align with previous studies that emphasize the detrimental effect of distribution shifts on model accuracy \cite{new_ood_1,new_ood_2,new_ood_3}, underscoring the importance of accounting for both inside and outside OOD scenarios in the development and evaluation of machine learning models.

\review{It is worth noting the conceptual similarities between our proposed "inside" and "outside" OOD framework and the established notions of internal and external validation prevalent in fields like medical research \cite{yang2024experts}. Internal validation, assessing model performance on data closely related to the training set, shares a resemblance with our "inside" OOD, where anomalies exist within the observed feature ranges. Conversely, external validation, evaluating generalizability to entirely new, independent datasets, echoes our "outside" OOD, where samples fall beyond the training data's feature boundaries. However, a key distinction lies in the basis of this categorization. While internal and external validation in medicine are often delineated by the source and context of the data (e.g., different patient cohorts or institutions), our "inside" and "outside" OOD are defined by a more direct, data-driven criterion based on whether the OOD samples fall within or beyond the observed range of each feature in the training data. This feature-centric definition allows for a more granular analysis of how different types of distributional shifts, both within and extending beyond the training data's span, uniquely impact machine learning model performance, highlighting a novel perspective on OOD analysis.}

The findings of this study provide two practical guidelines for practitioners. First, when developing and evaluating ML models, it is crucial to consider the potential presence of both inside and outside OOD data, as well as possible inside-outside profiles that can take place based on the available training data to predict possible OOD issues in production settings. Second, data augmentation techniques that simulate both inside and outside OOD conditions can help in training more resilient models \cite{last_2}. \review{For instance, inside OOD samples might be used to augment training data, thereby improving the robustness of models to unusual but plausible variations. In contrast, outside OOD samples could trigger more significant actions in a deployed system, such as initiating a model retraining process or activating safety protocols. Effectively, the inside-outside OOD framework offers a nuanced way to manage OOD, moving beyond a simple binary classification and enabling more adaptive and robust machine learning systems.}

\review{Nonetheless, this study is not without limitations. Mainly, the proposed OOD inside-outside profile assumes that the features are continuous. As such, categorical features are not addressed.} In addition, the synthetic dataset generation procedure is limited by the complexity and expressiveness of the datasets, which may not represent a portion of extremely complex real-world datasets. Future work should address these limitations to provide a more extensive understanding of OOD inside-outside profiling, which can be used to improve other properties of data-driven models, such as concept drift \cite{ds_ood_1,ds_ood_2}. 

\review{Taken jointly, the inside-outside OOD formalization facilitates a more robust and reliable machine learning development by recognizing and differentiating between these two types, allowing practitioners to better design targeted strategies to mitigate the adverse effects of OOD. The proposed framework opens several promising avenues for future investigation. First, exploring the interplay between inside-outside OOD and other related concepts, such as concept drift and adversarial examples, could provide valuable insights into the multifaceted challenges of distribution shift. Second, the development of practical tools and techniques for detecting and characterizing inside-outside OOD in real-world data would be of significant benefit to domains such as medical imaging, where the ability to reliably identify anomalous patterns is paramount, and autonomous systems, where robustness to unexpected scenarios is critical for safe operation \cite{tschuchnig2021anomaly}.}

\section*{Declarations}
\subsection*{Funding}
This study received no funding. 

\subsection*{Code avalability}
The code used for this study is freely available at: \url{https://github.com/teddy4445/inside_ood}. 

\subsection*{Conflicts of Interest/Competing Interests}
The author declares no conflict of interest. 

\subsection*{Acknowledgment}
The author wishes to thank both Oren Glickman and Assaf Shmuel for inspiring this study and Uri Itay for the thought-provoking discussion about it. 

\bibliography{biblio}
\bibliographystyle{unsrt}

\end{document}